\pdfoutput=1
\documentclass[11pt]{article}

\usepackage[]{acl}

\usepackage{times}
\usepackage{latexsym}

\usepackage[T1]{fontenc}
\usepackage[utf8]{inputenc}

\usepackage{microtype}

\usepackage{graphicx}

\title{Efficient Biomedical Entity Linking: Clinical Text Standardization with Low-Resource Techniques}

\author{Akshit Achara \thanks{\texttt{akshit.achara@gehealthcare.com}}  \hskip 2em Sanand Sasidharan  
	\thanks{\texttt{sanand.sasidharan@gehealthcare.com}} \hskip 2em
	Gagan N
	\thanks{\texttt{gagan.n@gehealthcare.com}}\\\\
	GE HealthCare
}

\usepackage{xcolor}
\usepackage{ulem}
\usepackage{bibentry}
\usepackage{amsmath}
\usepackage{amssymb}
\usepackage{subcaption, booktabs}
\usepackage{graphicx}
\usepackage{algorithm}
\usepackage{algorithmic}

\newcommand\algorithmicprocedure{\textbf{procedure}}
\newcommand{\algorithmicendprocedure}{\algorithmicend\ \algorithmicprocedure}
\makeatletter
\newcommand\PROCEDURE[3][default]{%
	\ALC@it
	\algorithmicprocedure\ \textsc{#2}(#3)%
	\ALC@com{#1}%
	\begin{ALC@prc}%
	}
	\newcommand\ENDPROCEDURE{%
	\end{ALC@prc}%
	\ifthenelse{\boolean{ALC@noend}}{}{%
		\ALC@it\algorithmicendprocedure
	}%
}
\newenvironment{ALC@prc}{\begin{ALC@g}}{\end{ALC@g}}

\begin{document}
\maketitle
\begin{abstract}
Clinical text is rich in information, with mentions of treatment, medication and anatomy among many other clinical terms. Multiple terms can refer to the same core concepts which can be referred as a clinical entity. Ontologies like the Unified Medical Language System (UMLS) are developed and maintained to store millions of clinical entities including the definitions, relations and other corresponding information. These ontologies are used for standardization of clinical text by normalizing varying surface forms of a clinical term through Biomedical entity linking.  With the introduction of transformer-based language models, there has been significant progress in Biomedical entity linking. In this work, we focus on learning through synonym pairs associated with the entities. As compared to the existing approaches, our approach significantly reduces the training data and resource consumption. Moreover, we propose a suite of context-based and context-less reranking techniques for performing the entity disambiguation. Overall, we achieve similar performance to the state-of-the-art zero-shot and distant supervised entity linking techniques on the Medmentions dataset, the largest annotated dataset on UMLS, without any domain-based training. Finally, we show that retrieval performance alone might not be sufficient as an evaluation metric and introduce an article level quantitative and qualitative analysis to reveal further insights on the performance of entity linking methods.
\end{abstract}

\section{Introduction and Related Work}
\label{introrel}
Medical text consists of a diverse vocabulary derived from various nomenclatures including varying surface forms corresponding to terms like diagnosis, treatment, medications, etc. This diversity poses a challenge for effective communication across medical institutions and organizations. One of the techniques to mitigate this inherent diversity present in multiple references to the same term is entity linking. Entity linking is used to map these references to standardized codes. These codes are curated and maintained by medical organizations for standardization of medical nomenclature. 

Given a corpus, \textit{entity linking} includes the mapping of a mention $m$ which is a span of $k$ words, to an entity $\epsilon$, where the entity belongs to a knowledge base such as Wikipedia. In the biomedical domain,  the textual phrases are linked with the corresponding concepts from a knowledge base constructed using the medical ontologies like UMLS~\cite{bodenreider2004unified}, SNOMED~\cite{el2018snomed}, etc. The UMLS ontology comprises  of a broad range of clinical entities along with rich information for each entity like synonyms, definitions, etc.  Traditional approaches for entity linking, such as Support Vector Machines~\cite{svm} and Random Forests~\cite{breiman2001random}, rely heavily on hand-crafted features, thereby restricting generalization to diverse data. Neural networks have emerged as a prominent technique for  entity linking due to their ability to learn semantic representations from textual data. 

Alias matching based techniques  like ~\cite{metamap,scispacy,sapbert}  have been proposed where an input mention is mapped to an alias associated with an entity in the knowledge-base. However, these techniques require large amount of training data. Contextualized entity linking approaches~\cite{krissbert}  utilize the semantic similarity between contextualized mentions. This approach requires a list of entities in advance and includes distant-supervision on articles containing examples of these entities. Generating medical codes using large language models can be error prone~\cite{soroush2024large}. In~\cite{yuan2022generative}, the authors use a seq2seq model to map a mention to its canonical entity name. This method is resource intensive and requires generation of synthetic examples for pretraining, utilizing entity definitions and synonyms. In~\cite{kong2021zero}, the authors propose a zero-shot entity linking approach by leveraging synonym and graph based tasks. However, the approaches require training samples from UMLS for both these tasks. Moreover, entity disambiguation has not been explored in the work.

Efficient student models like MiniLM~\cite{wang2020minilm} can be used to perform contrastive learning on synonyms of entities. This results in a significantly less embedding size ($384$) as compared to the approaches like SAPBERT~\cite{sapbert} with an embedding size of $768$.  The predicted candidates in alias based techniques are ranked based on the cosine similarity score. However, there are ambiguous cases where multiple entities have similar scores for a common mention. Therefore, there is a requirement to disambiguate these candidates through reranking. Cross-Attention based reranking approaches utilize supervised training on the concatenated mention and candidate representations as inputs~\cite{krissbert}.  More recent approaches utilize homonym disambiguation~\cite{garda2024belhd} and have shown to improve the performance of autoregressive approaches like GenBioEL.

In comparison to the discussed techniques, we propose an efficient and low resource zero-shot biomedical entity linking approach along with a suite of disambiguation techniques. Furthermore, we introduce an article level similarity analysis to obtain further insights. This also allows us to conduct a qualitative analysis without manually going through all the articles manually.

Our contributions are as follows:
\begin{enumerate}
	\item[\textbullet] \textbf{Data}:  We show that the impact of training is negligible on a finetuned MiniLM model\footnote{\url{https://huggingface.co/sentence-transformers/all-MiniLM-L6-v2}} as compared to the pretrained MiniLM model. Moreover, the pretrained MiniLM model when finetuned on all UMLS synonym pairs has worse performance than the all-MiniLM model.
	\item[\textbullet] \textbf{Disambiguation} We show that reranking on entity-level semantic information provided in UMLS can be highly effective for entity disambiguation. We further propose a parametric reranking technique that is beneficial for alias-based entity linking solutions.
	\item[\textbullet] \textbf{Evaluation}  We propose a comprehensive evaluation of entity linking which utilizes the semantic representation of articles coupled with the strict matching and related matching of predicted and gold standard entities. This evaluation is used to highlight issues related to the annotation granularity, missing context and surface form bias (for abbreviations) without the need of going through all the articles.
	
\end{enumerate}

\section{Datasets}
In this work, we explore entity linking on the Medmentions~\cite{mohan2019medmentions} dataset which consists of titles and abstracts from $4392$ English biomedical articles. These articles comprise of textual spans annotated with mentions of UMLS 2017AA entities. The dataset provides two versions:  a full version containing $34724$ unique entities and an st21pv version  with $25419$ unique entities, the latter being recommended by the authors for information retrieval.

\subsection{Preprocessing}
\label{preprocessing}
We replace the abbreviations with their corresponding full forms using Ab3p~\cite{sohn2008abbreviation}. The abbreviation expansion using Ab3p has shown to significantly improve the entity linking performance across different approaches~\cite{kartchner}. Prior to creating synonym pairs for training, we remove all the suppressed entities, deleted entities and deprecated entities. Some deprecated entities have also been merged with other entities having a synonymous relation. We map these deprecated entities  to the corresponding active entities with a synonymous relation.

\begin{table}[htpb!]
	\centering
	\begin{tabular}{lrr}
		\toprule
		&st21pv&full\\
		\midrule
		merged&181&280\\
		deleted&49&60\\
		non-synonymous&226&348\\		
		\bottomrule
	\end{tabular}
	\caption{The table shows the details of Medmentions entities annotated with  UMLS 2017AA version that are deprecated in UMLS 2023AB version. }
	\label{table:deprecated_2023AB}
\end{table}

Some annotations in Medmentions (prepared with UMLS 2017AA) are deprecated in the UMLS 2023AB (see details in table~\ref{table:deprecated_2023AB}). Therefore, approaches utilizing UMLS 2023AB version may want to use an updated version of Medmentions. Furthermore, the prototype space consisting of UMLS entities will have to be updated to the remove deprecated entities. This would help in avoiding deprecated entities to be predicted as candidates.

\section{Methodology}

In this work, we create a prototype vector space comprising of the encodings (feature vectors) associated with the canonical name of each entity in the UMLS ontology. To obtain meaningful encodings for constructing this prototype space, we train an encoder-based transformer~\cite{vaswani2017attention}  model on pairs of canonical names of entity synonyms. This is similar to the training approaches utilized in~\cite{kong2021zero} and ~\cite{sapbert}. The prototype space constructed using this trained model is used for performing semantic search, where the query encoding is obtained by passing the mention through the same model. This step is known as candidate generation. The candidate generation may lead to ambiguous results where multiple predicted entities have equal similarity scores. This is addressed through the reranking approaches discussed in section~\ref{rerankingmethodology}. Finally, we utilize both semantic similarity and retrieval performance for our quantitaive and qualitative evaluation. The comprehensive structure of our proposed approaches is depicted in the figure~\ref{fig:method}.

\begin{footnotesize}
	\begin{figure*}[htpb]
		\centering
		\includegraphics[width=0.95\textwidth, height=0.35\textwidth, keepaspectratio]{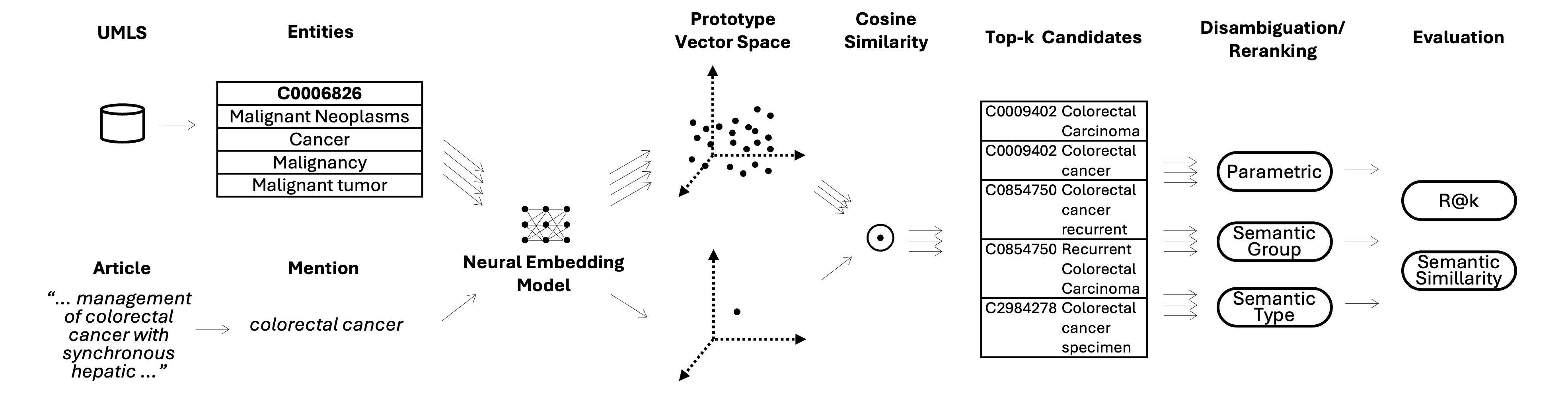}
		\caption{This figure illustrates the sequential flow of our proposed approaches. Starting from the left, we begin with leveraging a neural embedding model to create a prototype space on the UMLS entities. The cosine similarity metric is used to perform semantic search on the queries given the input mentions. The resultant top-$k$ candidates are reranked using the listed methods for disambiguation and finally a comprehensive evaluation comprising of the retrieval performance and semantic similarity is performed. }
		\label{fig:method}
	\end{figure*}
\end{footnotesize}

 The following sub-sections discuss the individual components used in our work:

\subsection{Training}
We construct a training dataset by taking all the canonical names for each entity from UMLS and create pairs of canonical names corresponding to the same entity. Each pair is of the form ($\epsilon_i$, $\epsilon^*_i$), where $\epsilon_i^*$ represents the canonical name of a synonym of entity $\epsilon_i$. The preprocessing steps are discussed in the section~\ref{preprocessing}. 
We use this dataset to finetune a sentence-transformer~\cite{reimers2019sentence} model using Multiple Negatives Ranking loss~\cite{henderson2017efficient}.  We use MiniLM~\cite{wang2020minilm} and a finetuned all-MiniLM\footnote{\url{https://huggingface.co/sentence-transformers/all-MiniLM-L6-v2}} model for training/finetuning on this dataset. The corresponding MiniLM and all-MiniLM models trained/finetuned on $k$ examples are hereafter referred as MiniEL$^*_k$ and  MiniEL$_{k}$ respectively.  For example, the all-MiniLM model finetuned on $10$ pairs/examples is referred as MiniEL$_{10}$. 

The Multiple Negatives Ranking loss function is defined as:
\begin{align}
	L(x, y, \theta) = \frac{1}{B} \sum_{j=1}^{B} \log P(y_j | x_j)
\end{align}

Here, $\theta$ represents the network parameters, $(x,y)$ represents a pair of phrases and $B$ represents the batch size. The parameters details for training are provided in section in Appendix in the section~\ref{terminologyparameters}.

\subsection{Candidate Generation}
A prototype space is prepared for the UMLS 2017AA version comprising of the encodings of canonical names of each entity and its synonyms. These encodings are computed using the MiniEL* and MiniEL models. The prototype space is used for performing semantic search where the queries are formed using the labeled mentions from the Medmentions dataset. The top-$k$ concepts are retrieved based on the cosine similarity of the query and entity encodings. These candidates are referred as generated candidates.

\subsection{Disambiguation}
\label{rerankingmethodology}

The candidate generation solely relies on the cosine similarity score between the mention and prototype space candidate encodings. However, there may be cases where multiple candidates have similar scores or the scores alone may not be sufficient to rank the candidates. Therefore, there is a need to rerank the candidates. We propose the following reranking approaches that to perform the entity disambiguation:

\subsubsection{Parametric Reranking}
\label{wosemantinf}

In this section, we propose a parametric approach to rerank the generated candidates. We consider three parameters based on the prototype space and our training framework for disambiguation namely, cosine similarity score (CSS), representative alias score (RAS) and the candidate entity frequency score (CEFS) as our parameters. The parameters have the corresponding coefficients $a$, $b$ and $c$ respectively.  These parameters are used to compute a new ranking score for each candidate. The equation below shows the updated score ($\delta^*(.,.)$) computation for reranking each of the generated candidates.

\begin{footnotesize}
\begin{align}
	\delta^*(q, v) = a*\delta(q,v) + b*\frac{1}{n}\sum_{j=1}^{n} \delta(q,v_j) + c*n
\end{align}
\end{footnotesize}

Here, $q$ is the query encoding, $v$ is a generated candidate encoding and $n$ is the number of aliases of $v$ in the generated candidates.

The optimal selection of coefficients $a$, $b$ and $c$ corresponding to each of these parameters is performed through a grid search on a subset of manually defined bounds. Further details on the grid search and the impact of the these coefficients are discussed in appendix in the section~\ref{parametericablation}. 

\subsubsection{With UMLS Semantic Information}
\label{semantinf}
UMLS consists of additional classification associated with individual entities, grouping them based on their semantic types and semantic groups. Each semantic type and semantic group has a canonical name. In this section, we calculate the cosine similarity between the mention's semantic type or semantic group canonical name encoding and the corresponding canonical names of the top-$k$ candidates. This similarity score is added to the initial candidate generation score to rerank the top-$k$ candidates.

\begin{enumerate}
	\item  \textbf{Assuming Availability of Gold Standard Information}:
	In this case, we assume that the gold standard semantic type and semantic group information is available for each mention.  We rerank the candidates by utilizing the following methods:
	\begin{enumerate}
		\item \textbf{Semantic Type Based Disambiguation}: In this method, calculate the cosine similarity between canonical name 
		encodings of semantic types of a
		mention and each of its top-$k$ candidates. The updated score is computed as follows: 
		
		\begin{footnotesize}
				\begin{align}
				\centering
				\delta^*(q, v) = \delta(q, v) + \delta(TUI(q), TUI(v))
			\end{align}
		\end{footnotesize}
		Here, $TUI(.)$ maps the input mention/entity to the encoding of corresponding semantic type canonical names.
		\item  \textbf{Semantic Group Based Disambiguation}: In this method, calculate the cosine similarity between canonical name 
		encodings of semantic groups of a
		mention and each of its top-$k$ candidates. The updated score is computed as follows: 
		
		\begin{footnotesize}
		\begin{align}
			\delta^*(q, v) = \delta(q, v) + \delta(SG(q), SG(v))
		\end{align}
		\end{footnotesize}
		Here, SG(.) maps the input mention/entity to the encoding of the corresponding semantic group canonical names.
	\end{enumerate}
	\item \textbf{Semantic Type/Group Prediction}: In scenarios where the semantic type/group information of the mentions is not available, the methods proposed in~\cite{tuidisambclassification} and~\cite{twodisamb} can be used to predict the semantic type or group based on the input mentions. This can be followed by the computational methods discussed in the section~\ref{semantinf}.
	
\end{enumerate}

\section{Results and Discussion}
\label{results}

We obtain the retrieval performance for the discussed approaches by  considering the top-$k$ closest candidates (that include aliases) from the prototype space.  We observe that the retrieval performance (considering top-$128$ candidates) of all the miniEL and miniEL$_{1000}$ approaches is around $87\%$ for the st21pv version and $88\%$ for the full version of the Medmentions dataset.

\subsection{Quantitative Analysis}

In this section, we present the quantitive analysis associated with candidate generation (see section~\ref{dataexperiments} and tables~\ref{table:dataperfst21pv},~\ref{table:dataperffull}) and reranking (see section~\ref{disambperf}, figure~\ref{fig:disambtrends} and tables~\ref{table:disambresults}). Furthermore, the intricate analysis on the distribution of exact, related and missed candidate matches are discussed in the section~\ref{reldiscussion}.

\subsubsection{How much data do we need?}
\label{dataexperiments}

In this section, we discuss the candidate generation performance of our approaches trained using varying number of examples. It can be seen that the performance of miniEL has a negligible training impact and the performance is stable across different number of examples (see tables~\ref{table:dataperfst21pv} and~\ref{table:dataperffull}).  However, the miniEL$^*$ approach improves consistently with increasing number of training examples.The miniEL approach without any finetuning still outperforms the miniEL$^*$ approach trained on all the training examples. 

\begin{table}[htpb!]
	\centering
	\resizebox{.45\textwidth}{!}{% <------ Don't forget this %
	\begin{tabular}{l|rr|rr}
		\toprule 
		& \multicolumn{2}{c|}{miniEL*} & \multicolumn{2}{c}{miniEL} \\
		\midrule
		Training Samples & R@1 & R@5 & R@1 & R@5 \\
		\midrule
		0&0.401&0.594&0.553&0.756\\
		10 &0.427 &0.622&0.552&0.758\\
		%100 &0.451  &0.647&0.550 &0.759\\
		1000 &0.499 &0.693&0.557&0.766\\
		%5000 &0.519 &0.714&0.550&0.761\\
		10000 &0.518&0.717 &0.553&0.76\\
		ALL &0.534 &0.736&0.556&0.756\\
		\bottomrule
	\end{tabular}% <------ Don't forget this %
}
	\caption{This table shows the R@1 and R@5 candidate generation performance of the approaches on the Medmentions (st21pv) dataset. The models are trained with varying number of training samples used to train/finetune the MiniEL$^*$ and MiniEL models.}
	\label{table:dataperfst21pv}
\end{table}

\begin{table}[htpb!]
	\centering
	\resizebox{.4\textwidth}{!}{% <------ Don't forget this %
	\begin{tabular}{l|rr|rr}
		\toprule 
		& \multicolumn{2}{c|}{miniEL*} & \multicolumn{2}{c}{miniEL} \\
		\midrule
		Training Samples & R@1 & R@5 & R@1 & R@5 \\
		\midrule
		0&0.462&0.657&0.567&0.782\\
		10 &0.477 &0.676&0.565 &0.783\\
		%100 &0.490  &0.690&0.564 &0.783\\
		1000 &0.525 &0.728&0.569&0.789\\
		%5000 &0.539 &0.746&0.567&0.787\\
		10000 &0.537 &0.747&0.568&0.788\\
		ALL &0.556 &0.761&0.568&0.783\\
		\bottomrule
	\end{tabular}% <------ Don't forget this %
}
	\caption{This table shows the R@1 and R@5 candidate generation performance of the approaches on the Medmentions (full) dataset. The models are trained with varying number of training samples used to train/finetune the MiniEL$^*$ and MiniEL models.}
	\label{table:dataperffull}
\end{table}

\subsubsection{Reranking Performance}
\label{disambperf}
In the following subsections, we discuss the candidate reranking results. The results corresponding to the parametric approach and those corresponding to the semantic disambiguation approaches are discussed in the following subsections.

\begin{figure}[htpb]
	\centering
	\resizebox{8cm}{!}{\includegraphics[width=\textwidth, height=0.8\textwidth, keepaspectratio]{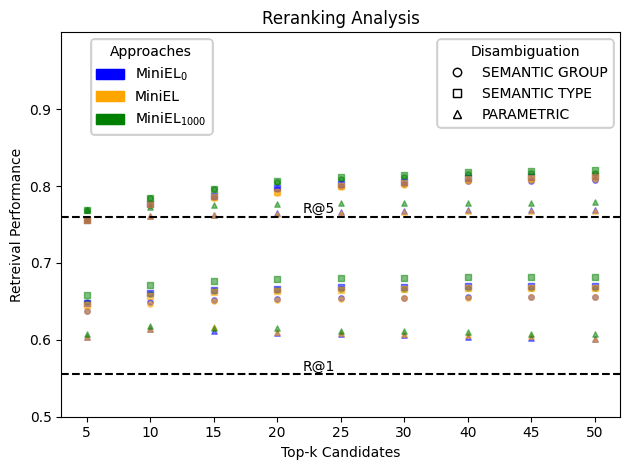}}
	\caption{This figure highlights the trends associated with the retrieval performance improvement over varying top-k candidates using MiniEL$_0$, MinEL and MiniEL$_{1000}$ models. The improvement in R@1 is more significant as compared to that in R@5 for all the models and reranking methods. It can be observed that the retrieval performance of \textit{PARAMETRIC} reranking decreases with increase in the top-$k$ (k>15) whereas the performance of \textit{SEMANTIC GROUP} and \textit{SEMANTIC TYPE} reranking is consistent across the top-$k$.}
	\label{fig:disambtrends}
\end{figure}

\begin{enumerate}
	\item \textbf{Parametric Reranking}:
	The top-$k$ candidates selected based on the parametric approach discussed in section~\ref{wosemantinf} and the corresponding results are shown in  figure~\ref{fig:disambtrends} and table~\ref{table:disambresults}. It can be seen that the retrieval performance improves by from $0.553$ to $0.614$ for the st21pv version and from $0.567$ to $0.630$ for the full version of Medmentions. The $a$, $b$ and $c$ values used to obtain these results have the proportion $a:b:c\propto50:2:1$ (see section~\ref{parametericablation} for more details).
	
	\item \textbf{With UMLS Semantic Information}:
	In this section, we discuss the retrieval performance improvements after the reranking using the semantic type and group information. The details of these methods are discussed in section~\ref{semantinf}.
	
	\begin{table}
		\centering
		\resizebox{.45\textwidth}{!}{% <------ Don't forget this %
		\begin{tabular}{l|rr|rr}
			\toprule 
			& \multicolumn{2}{c|}{st21pv} & \multicolumn{2}{c}{full} \\
			\midrule
			Reranking&top-5&top-10&top-5&top-10\\
			\midrule
			PARAMETRIC&0.604&0.614&0.620&0.630\\
			GROUP&0.638&0.649&0.659&0.670\\
			TYPE&0.648&0.661&0.681&0.697\\
			\bottomrule
		\end{tabular}% <------ Don't forget this %
	}
		\caption{The table shows the R@1 performance of the MiniEL$_0$ model after applying the listed reranking methods using the top-$5$ and top-$10$ candidates.}
		\label{table:disambresults}
	\end{table}

	 Figure~\ref{fig:disambtrends} and table~\ref{table:disambresults} show the R@1 performance of the MiniEL$_0$ model after applying these reranking strategies. The performance improves from $0.553$ to $0.649$ for semantic group and to $0.661$ for semantic type reranking for the st21pv version of Medmentions. Similar observations can be made for the full version of Medmentions.  Moreover, the retrieval performance does not deviate significantly with the increase in the top-k candidates used for reranking (see figure~\ref{fig:disambtrends} for details).
\end{enumerate}

The improvement in candidate ranking is approached in two ways. Firstly, to maximize the R@1 performance by reranking the generated candidates (see details in section~\ref{rerankingmethodology}) and secondly, to include context for addressing the context based ambiguity (see details in Appendix in section~\ref{contextdisamb}).

\subsubsection{How should the performance be evaluated?}
\label{reldiscussion}
In the retrieval-based evaluation strategy, we compute the retrieval performance on gold standard and predicted entity matches. However, there are cases where the most similar candidate is related to the gold standard entity. It can be seen in the table~\ref{table:relationandperformance} that about $77\%$ entities are exacting matching or are related to the gold standard entity. The details of each type of relation we have considered are provided by UMLS.\footnote{\url{https://www.nlm.nih.gov/research/umls/knowledge_sources/metathesaurus/release/abbreviations.html\#mrdoc\_REL}}

\begin{table}[htpb]
	\resizebox{.45\textwidth}{!}{% <------ Don't forget this %
		\begin{tabular}{lrrr}
			\toprule
			Approach&Exact&Related&Missed\\
			\midrule
			MiniEL$_0$&0.553&0.220&0.227\\
			MiniEL$_0$ + PARAMETRIC&0.614&0.172&0.214\\
			MiniEL$_0$ + GROUP&0.649&0.188&0.163\\
			MiniEL$_0$ + TYPE&0.661&0.176&0.163\\
			\bottomrule
		\end{tabular}% <------ Don't forget this %
	}
	\caption{This table shows the $R@1$ retrieval performance distributed into the exact matches, related matches and missed matches. The top-$10$ candidates are used for reranking. Here, we use the st21pv version of Medmentions.}
	\label{table:relationandperformance}
\end{table}

\begin{figure}[htpb]
	\centering
	\resizebox{7.7cm}{!}{\includegraphics[width=\textwidth, height=0.65\textwidth, keepaspectratio]{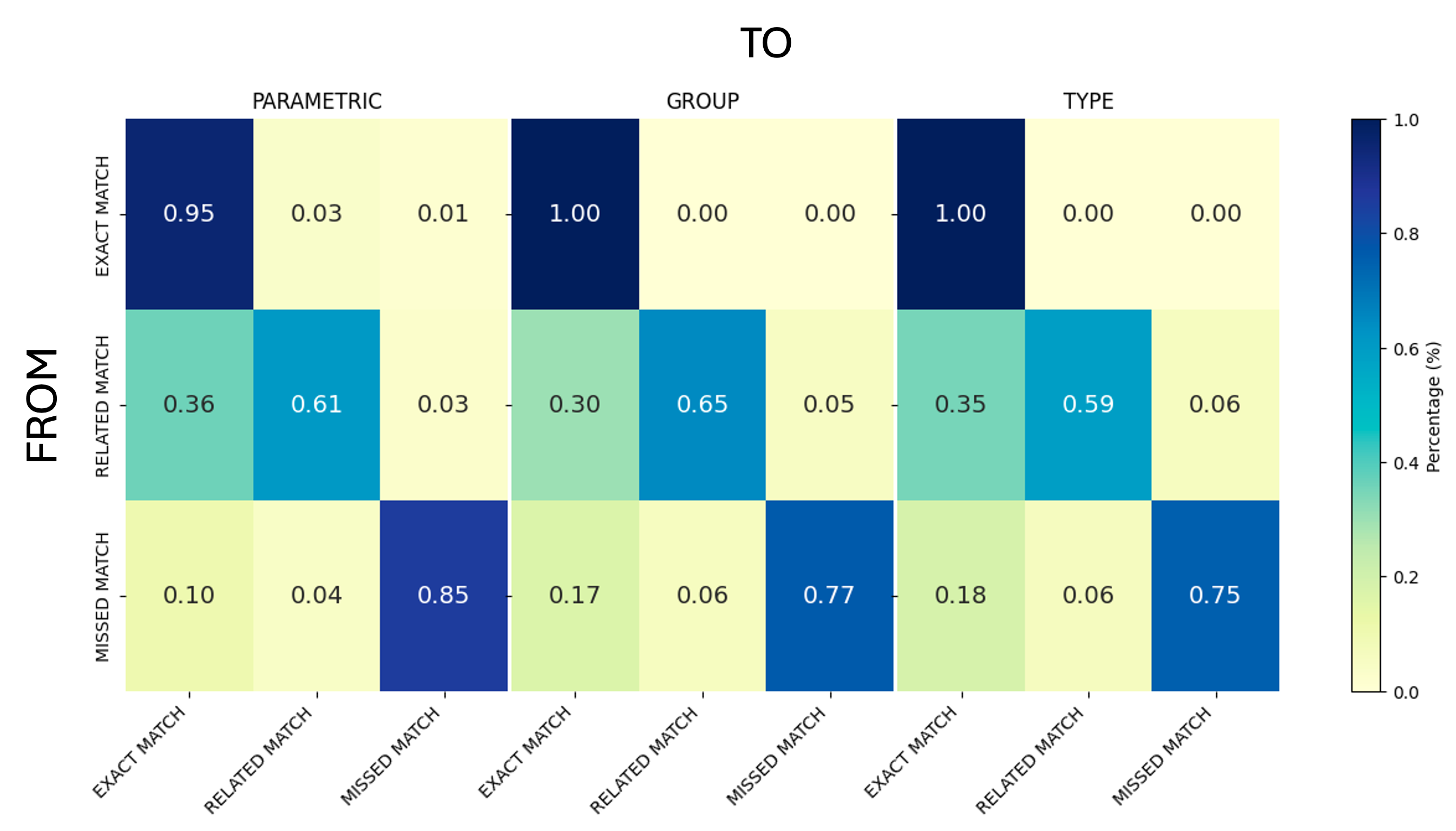}}
	\caption{This heatmap illustrates the percentage changes in the number of initial exact, related and missed matches for the MiniEL$_0$ model. The performance preceding the changes is labeled 'FROM' for the rows, while the subsequent performance is denoted by 'TO' for the columns. The experiments are performed on the st21pv version of Medmentions.}
	\label{fig:matchchanges}
\end{figure}

Figure~\ref{fig:matchchanges} shows that the effect of parametric reranking is directed primarily towards converting related matches to exact matches, coverting $36\%$ of related matches into exact matches. The semantic group and semantic type based reranking approaches convert both missed and related matches into exact matches.

\begin{figure}[htpb]
	\centering
	\resizebox{7.7cm}{!}{\includegraphics[width=\textwidth, height=0.8\textwidth, keepaspectratio]{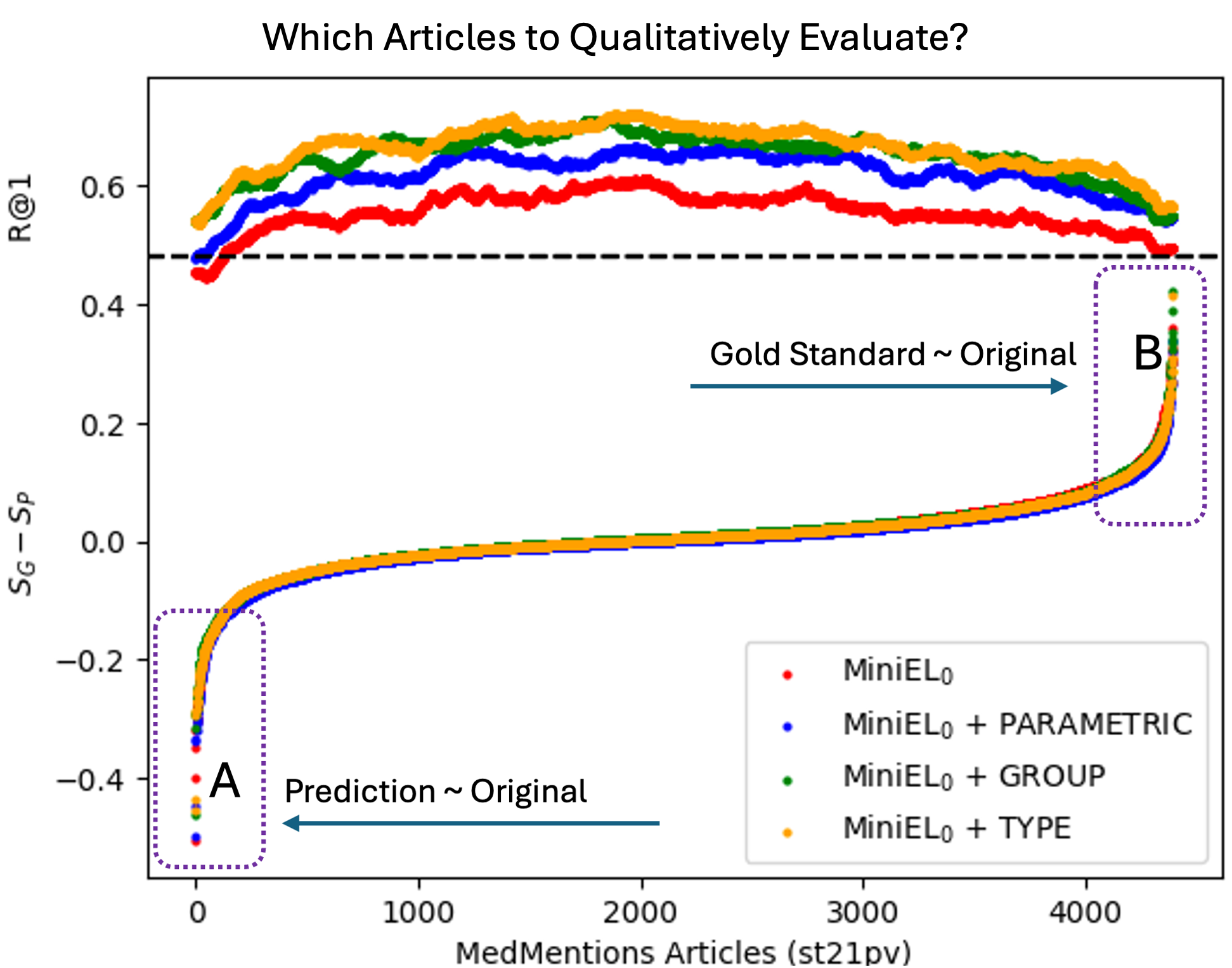}}
	\caption{This figure illustrates the disparity in similarity scores ($S_G - S_P$) at the article level (4392 articles), alongside the smoothed retrieval performance (R@1) per article using a moving average with a window size of 200. The region $A$ consists of semantically closer predictions and $B$ consists of semantically farther predictions.}
	\label{fig:retsim}
\end{figure}

The following analysis is focused on the further evaluation of related and missed matches. In this article level analysis,  we replace a mention with the closest generated candidate's canonical name for each mention in the article where the closest candidate is a related match or a missed match respectively. This results in an article $A_P$. We compute the cosine similarity between the original article $A$ and the modified aricle $A_P$ called $S_P$ using a PubmedBERT-base~\cite{pubmedbert} model\footnote{\url{https://huggingface.co/NeuML/pubmedbert-base-embeddings}} finetuned using sentence transformers~\cite{reimers2019sentence} on biomedical data.  Similarly, we also replace the mentions with the gold standard canonical names to create an article $A_G$. This is followed by computation of cosine similarity between $A$ and $A_G$ called $S_G$.  We focus on scenarios where $S_P$ and $S_G$ deviate significantly as compared to the mean deviation of the articles. These are highlighed in the figure~\ref{fig:retsim}. This forms a base for our qualitative analysis where we use this deviation to provide insights on the granularity of gold standard predictions as well as highlight current issues in the approach. 

\subsection{Qualitative Analysis}
 \label{qualitative}
 We perform a qualitative analysis on the entity linking predictions to highlight the difference in the granularity of the gold standard and predicted entities.
 
 In this section, we qualitatively evaluate the articles displayed in the regions $A$ and $B$ of figure~\ref{fig:retsim}. The region $A$ consists of articles where the predicted article $A_P$ is semantically more similar to the original article $A$ as compared to the gold standard article$A_G$. Whereas, the region $B$ consists of articles where $A_G$ is more similar to $A$ as compared to $A_P$. 
 
 Table~\ref{table:regionAqualitative} shows the qualitiative examples from region A where it can be observed that our approach is penalized for granular or highly related predictions. For example, The mention \textit{gene expression classifier} has a gold standard entity \textit{\uline{Research Activities}} as compared to the more granular prediction \textit{\uline{Gene Expression Profiling}}. Similarly, the mention \textit{cytoplasmic tails} has a gold standard enttity \textit{\uline{CytoPlasmic}} as compared to the more granular prediction \textit{\uline{Cytoplasmic Domain}}. 
 
 \begin{table}[t]
 	\resizebox{7cm}{!}{\begin{tabular}{p{10cm}}%
 		\toprule
 	\textbf{\texttt{MENTION}}: \textit{"\uline{Vitamin D Receptor Activator} Use and Cause-specific...\uline{Vitamin D receptor activators} (VDRA) may exert...5,635 \uline{VDRA} users were matched...that \uline{VDRA} use was"}\\
	\textbf{\texttt{GOLD}}: \textcolor{blue}{Biologically Active Substance} (C0574031)\\
	\textbf{\texttt{PREDICTION}}: \textcolor{blue}{VDR protein, human} (C3657722) with parent entity \textcolor{brown}{Vitamin D3 Receptor} (C0108082) \\
	\midrule
	\textbf{\texttt{MENTION}}: \textit{"Influence of \uline{Sinus} Floor Configuration....the \uline{sinus} floor configuration...osteotome \uline{sinus} grafting procedure...into the \uline{sinus area}...\uline{sinus} floor configuration...\uline{sinus} floor profile...flat \uline{sinus} group...maxillary \uline{sinus} following...predictable in \uline{sinuses} with a concave..."}\\
	\textbf{\texttt{GOLD}}: \textcolor{blue}{Anatomical space structure} (C0229984)\\
	\textbf{\texttt{PREDICTION}}: \textcolor{blue}{Nasal sinus} (C0030471)\\
	\midrule
	\textbf{\texttt{MENTION}}: \textit{"...effectiveness of \uline{disc synoptoscope} on patients...effectiveness of \uline{disc synoptoscope} on binocularity...therapy with \uline{disc synoptoscope} in...with \uline{disc synoptoscope} is effective...\uline{disc synoptoscope} could serve as an..."}\\
	\textbf{\texttt{GOLD}}: \textcolor{blue}{Medical Devices} (C0025080)\\
	\textbf{\texttt{PREDICTION}}: \textcolor{blue}{Synoptophores}  (C0183765)\\
	\midrule
	\textbf{\texttt{MENTION}}: \textit{"...performance of the \uline{Afirma gene expression classifier}...the Afirma \uline{gene expression classifier} (GEC)...on which \uline{GEC} was performed...\uline{GEC} testing was performed...\uline{atypia of undetermined significance} (AUS)...the \uline{AUS} cases...the \uline{AUS} group...patients with \uline{AUS}...value of \uline{GEC} decreased from...suspicious \uline{GEC} result...value of \uline{GEC} in indeterminate...suspicious \uline{GEC} result...suspicious \uline{GEC} result..."}\\
	\textbf{\texttt{GOLD}}:  \textcolor{blue}{Research Activities} (C0243095), \textcolor{blue}{Finding} (C0242481)\\
	\textbf{\texttt{PREDICTION}}: \textcolor{blue}{Gene Expression Profiling} (C0752248), \textcolor{blue}{Atypical cells of undetermined significance}  (C0522580)\\
	\midrule
 	 %\textbf{\texttt{MENTION}}: \textit{Nanosponge -Based \uline{ Multistimuli-Responsive Drug Vehicles} for Targeted \uline{ Chemo-Photothermal Therapy}}\\
 	%\textbf{\texttt{GOLD}}: \textcolor{blue}{Drug vehicle} (C0042444), \textcolor{blue}{Therapeutic procedure} ( C2930750)\\
 	%\textbf{\texttt{PREDICTION}}: \textcolor{blue}{\uline{Pharmaceutical Vehicles}} (C2930750), \textcolor{blue}{\uline{Photochemotherapy}} (C0031740)\\
 	%\midrule
 	 \textbf{\texttt{MENTION}}: \textit{"including the  \uline{cytoplasmic tails} of integrins and components of the actin cytoskeleton"}\\
 	\textbf{\texttt{GOLD}}: \textcolor{blue}{CytoPlasmic} (C0521449)\\
 	\textbf{\texttt{PREDICTION}}: \textcolor{blue}{\uline{Cytoplasmic Domain}} (C1511625) with alias 'Cytoplasmic Tail'. \\
 	\bottomrule
 \end{tabular}%
}
\caption{The table shows qualitative examples selected from the region A in the figure~\ref{fig:retsim}.}
\label{table:regionAqualitative}
 \end{table}
 
  Table~\ref{table:regionBqualitative} shows the qualitatuve examples corresponding to the region B where it can be seen that the gold standard annotation is based on the context of mention in the article. More specifically, the mention \textit{mice} has a gold standard emtity: \textit{\uline{Laboratory mice}} based on the article context. However, this context is missing in the mention surface form. Therefore, to address these kind of  cases, we need to provide the necessary context in the query. We utilize three different disambiguation techniques and show examples of the corresponding predictions. We observe that additional context from the articles may result in granular predictions. However, the results are highly sensitive to the context and overall retrieval performance drops significantly (see section~\ref{contextdisamb} for more details).
  
  We also observe an inconsistency in the granularity of gold standard entities in these examples. The mention \textit{experimental mice} has a gold standard entity \textit{\uline{Animals, Laboratory}} as compared to the more granular prediction \textit{\uline{Laboratory mice}}.
  
 \begin{table}[htpb!]
	\resizebox{7cm}{!}{	\begin{tabular}{p{10cm}}%
	\toprule
  	 \textbf{\texttt{MENTION}}: \textit{"....iron accumulation in the substantia nigra (SN) of \uline{mice}.....the substantia nigra of \uline{experimental mice}  treated with MPTP}."\\
  	\textbf{\texttt{GOLD}}: \textcolor{blue}{Laboratory mice} (C0025929), \textcolor{blue}{Animals, Laboratory} (C0003064)\\
  	\textbf{\texttt{PREDICTION}}: \textcolor{blue}{House mice} (C0025914), \textcolor{blue}{Laboratory mice} (C0025929)\\  	
  	\midrule
  	\textbf{\texttt{MENTION}}: \textit{"...mRNA N6-methyladenosine \uline{methylation} of postnatal...mRNA m6A \uline{methylation} during...outcomes of mRNA m6A \uline{methylation}...levels of m6A \uline{methylation} and...by m6A \uline{methylation} at...higher m6A \uline{methylation} and...differential m6A \uline{methylation} may..."}\\
  	\textbf{\texttt{GOLD}}: \textcolor{blue}{mRNA methylation} (C2611689)\\
  	\textbf{\texttt{PREDICTION}}: \textcolor{blue}{Methylation} (C0025723)\\
  	\bottomrule
 \end{tabular}%
}
\caption{The table shows qualitative examples selected from the region B in the figure~\ref{fig:retsim}.}
\label{table:regionBqualitative}
\end{table}

\section{Conclusion}

Biomedical entity linking has been an active area of research with various approaches being proposed to improve medical text standardization (see details in section~\ref{introrel}). We propose a multi-stage approach where the first stage retrieves candidates with a high recall ($\sim87\%$ for top-$128$ candidates). This is followed by application of the proposed reranking approaches focused on improving the R@1 retrieval performance. The reranking improves the performance by more than $10\%$ (see figure~\ref{fig:disambtrends} and table~\ref{table:disambresults}).  We investigate the misses in R@1 and segregate the candidates into related and missed matches. Following this, we compute the article level semantic similarity together with the article level retrieval performance. This analysis highlights qualitative examples that can be used to obtain further insights about the framework. The semantic analysis is used to select the following types of qualitative examples: a) low retrieval performance and high similarity and, b) low retrieval performance and low similarity. The former can be highlight issues pertaining to granularity of gold standard entities and the latter can be used to highlight issues pertaining to the retrieval performance. Overall, the proposed techniques are highly effective in entity linking and have negligible training, prototype-space creation and inference costs (see table~\ref{table:complexity} for more details).

\subsection{Future Scope}

We believe that there is a significant scope for future developments in biomedical entity linking across different components of existing deep learning solutions. Firstly, there can be multiple biomedical normalizations for a mention or surface form. However, there is no method to determine the "closeness" of a prediction to a surface form as opposed to the binary matching. We believe that there should be a partial scoring instead of a binarized computation in order to accomodate the quality of predictions in the evaluation.  Moreover, semantic similarities can also determined by experts to provide a ranking that could be used across biomedical entity linking for disambiguation.

\subsection{Limitations}
We observe that while an abbreviation pre-processing module is utilized in the proposed approaches, it doesn't convert all the abbreviations into their full forms. This causes a high amount of ambiguity in the results and often times the retrieval candidates do not consist of the correct entity. This drawback in positive pairs based learning has also been highlighted in~\cite{krissbert}. Research addressed towards improving abbreviation expansion can help improve the recall of our candidate generation.  Moreover, the region $B$ in figure~\ref{fig:retsim} highlights the examples where missing context in the surface form causes our framework to predict broader entities as the closest candidates. We utilize various approaches to include additional implicit and explicit context into our queries and analyze the corresponding retrieval performance (see details in Appendix section~\ref{contextdisamb}).

\bibliography{custom}
\newpage
\appendix
\section{Appendices}
\subsection{Terminology and Parameters}
\label{terminologyparameters}
This section includes the terminology details and training, inference or other parameters used in this work.

\begin{table}[htpb!]
	\centering
	\resizebox{7cm}{!}{\begin{tabular}{lp{9cm}}%
		\toprule
		Term&Description\\
		\midrule
		$\delta$& Similarity function\\
		\hline
		$m$& mention\\
		\hline
		$\epsilon$ & Entity\\
		\hline
		$q$ &Query\\
		\hline
		$\mu$&Entity canonical name\\
		\hline
		$TUI(.)$&\shortstack{maps an entity to it's semantic type canonical name}\\
		\hline
		$SG(.)$&\shortstack{maps an entity to it's semantic group canonical name}\\
		 \hline \\[0.01cm]
		$R@n$&\shortstack{Retrieval performance on top-$n$ unique candidate entities}\\
		\hline
		top-$k$& \shortstack{top-$k$ candidate entities including aliases}\\
		\bottomrule
	\end{tabular}%
}
	\caption{The table shows the symbols used in our work and the corresponding descriptions.}
	\label{table:terminology}
\end{table}

Table~\ref{table:complexity} shows the memory consumption and carbon emissions associated with the MiniEL$_0$ approach. It can be seen that our proposed techniques is low resource and results in very low amount of carbon emissions.

\begin{table}[htpb!]
	\begin{tabular}{p{2.45cm}p{1.3cm}p{2.3cm}}
		\toprule
		Phase&Memory (MB) &Emissions (Kg. Eq. CO2)\\
		\midrule
		Training&0&0\\
		Prototype Space Creation&1906&0.1\\
		Inference&938&0.04\\
		\bottomrule
	\end{tabular}
	\caption{The table shows the memory and carbon emission details. We utilized a 16GB V100 GPU for our tasks. The Inference was performed on the st21pv version of Medmentions.}
	\label{table:complexity}
\end{table}

\subsection{Ablation Studies}
\label{parametericablation}
In this section, we discuss the influence of parameters used in the parametric disambiguation approach discussed in the section~\ref{rerankingmethodology}. Specifically, we consider the candidate generation results obtained by using the MiniEL$_0$ model and perform the reranking by removing $b$ and $c$ parameters respectively. To highlight the impact of changing the $a$, $b$ and $c$ values, we perform a grid search on a manually selected range of values.
\begin{figure}[htpb]
	\centering
	\resizebox{7.5cm}{!}{\includegraphics[width=\textwidth, height=0.9\textwidth, keepaspectratio]{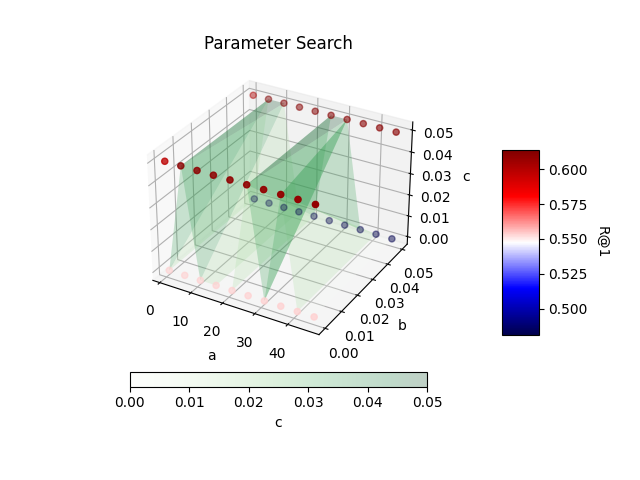}}
	\caption{This figure shows the grid search on the parameters $a$, $b$ and $c$ for optimizing the R@1 performance of the MiniEL$_0$ model using the parametric approach discussed in the section~\ref{rerankingmethodology}. The optimal combination of $a$, $b$ and $c$ is found to be $5$, $0.1$ and $0.05$, respectively.}
	\label{fig:gridsearch}
\end{figure}

Furthermore, considering the top-$10$ candidates for reranking, removal of  the parameter $b$ results in an R@1 of $0.611$, removal of $c$ results in $0.481$. This can be compared to the baseline R@1 $0.553$ and the R@1 of $0.614$ obtained using optimal $a$,$b$ and $c$. The performance is computed on the st21pv version of Medmentions. Overall, the impact of parameter $c$ is highly significant in the performance improvement.

\subsection{Contextualized Queries}
\label{contextdisamb}
In our framework, the encoded representations of mentions are queried on the prototype space to get relevant candidates from UMLS. However, the mention spans alone may lack the necessary context to map the mention to their corresponding UMLS entities. In this section, we evaluate multiple techniques for incorporating context in the queries. Specifically, we use a running span based context addition, an implicit context addition and an attention based span context addition.

\subsubsection{Neighboring Context}

In this approach, we select a few words before and after the mention span to update the mention $m$ and encode the updated mention to form a query. 

Firstly, we add $5$ neighbouring words before and after the mention and observe that the retrieval performance drops drastically ($R@1 \sim 10\%$). Therefore, we the number of words to $2$ on both sides of the mention which results in a drastic drop in retrieval performance  ($R@1 \sim 22\%$).

Overall, this context addition approach results in a significant drop in our retrieval performance and may not be suitable for contextual disambiguation.

\subsubsection{Attention-Based Context}
\label{attentionspanenrichment}

In this section, we perform experiments to identify the most influential words from the articles that attend to the span in consideration. We modify the original mentions by adding these words as additional context. This is done by utilizing the attention mechanism of encoder based transformer models namely BioBERT~\cite{lee2020biobert}. Firstly, the entire title and abstract text is tokenized and passed to these models. The corresponding attention outputs are obtained and passed to the mention enrichment algorithm. 

Let $k$ be the number of word-piece tokens obtained from the encoder model, for each head $H$ of Layer $L$, the attention matrix can be mentioned as:

\begin{footnotesize}
\begin{align}
	A = \begin{bmatrix}
		a_{11}& a_{12}& ...&a_{1k}\\
		a_{21}& a_{22}& ...&a_{2k}\\
		\vdots& \vdots& ....& \vdots\\
		\vdots& \vdots& ....& \vdots\\
		a_{k1}& a_{k2}& ....& a_{kk}
	\end{bmatrix}
\end{align}
\end{footnotesize}

The mention spans lie in the range $[c, d]$ where $0 \leq c < d \leq  k$. Therefore, the matrix $A$ can be shortened to a submatrix of interest $B$ mentioned as:

\begin{footnotesize}
\begin{align}
	B = \begin{bmatrix}
		a_{cc}& a_{c(c+1)}& ...&a_{cd}\\
		a_{(c+1)c}& a_{(c+1)(c+1)}& ...&a_{(c+1)d}\\
		\vdots& \vdots& ....& \vdots\\
		\vdots& \vdots& ....& \vdots\\
		a_{kc}& a_{k(c+1)}& ....& a_{kd}
	\end{bmatrix}
\end{align}
\end{footnotesize}
Equivalently,
\begin{align}
	B = \begin{bmatrix}
		b_c& b_{(c+1)}& ...&b_d\\
	\end{bmatrix}
\end{align}
where $b_{i}$ represents a column of $B$. Next, the token corresponding to the maximum attention value of each column is obtained as $T(max(b_i))$ where $T(j)$ represents the token at index $j  \in \{1,2,...,k\}$in the text spanning from $1^{st}$ to the $k^{th}$ token.  The resulting token vector from the attention head $H_m$ and Layer $L_n$ is represented as:

\begin{footnotesize}
	\begin{align}
	R_{nm} = \begin{bmatrix}
		T(max(b_c)& T(max(b_{(c+1)})& ...&T(max(b_d)\\
	\end{bmatrix}
\end{align}
\end{footnotesize}

The \textit{ENRICH} function discussed in the algorithm~\ref{alg:algorithm-label} return the enriched context for a given mention $m$,  which is then modified as shown below:
\begin{align}
	m* = m : R_{mn}[0], R_{mn}[1]
\end{align}

Finally, stop words are removed from $R_{mn}[0]$ and $R_{mn}[1]$.
An example mention cold can be modified as \textit{cold: \textbf{severe,recent}} where, 'severe, recent' is the added context.

\begin{algorithm}
	\caption{Enrichment Context Selection}
	\label{alg:algorithm-label}
	\begin{footnotesize}
	\begin{algorithmic}
		\PROCEDURE[most common by length in descending order]{sort$_{mcbl}$}{$V\colon \mathrm{1D~vector}$}
		\STATE $ C = \{x~|~count(x) = max(count(T))~\forall~T \in V\}$
		\STATE $C^* = \{x~|~x \in C~and~ len(x) >= len(y)~\forall~y \in C\}$
		\RETURN $C^{*}$ \\
		\ENDPROCEDURE\\
		\COMMENT{$R_{n}$ denotes the representative token from all attention heads in Layer n}\\
		\COMMENT{$L_{n}$ denotes the representative token(s) from Layer n}\\
		\COMMENT{$M$ denotes the representative token(s) for the token$_k$ in mention M}\\
		\COMMENT{$E$ denotes the representative token context (E) for mention M}
		\PROCEDURE[enrich mention with context]{enrich}{$R_n\colon \mathrm{1D~vector}$}
		\STATE $C^* = SORT_{mcbl}(R_{nm})$
		\STATE $R_{n} = C^*_1$ or $R_{n} = C^*(1)$
		\STATE $L_n = \{R_1, R_2, ..., R_z\}$
		\STATE $C^* = SORT_{mcbl}(L_{n})$
		\STATE $M_t = \{C^*_1, C^*_2\}$
		\STATE $M = \{M_1, M_2, ..., M_k\}$
		\STATE $C^* = SORT_{mcbl}(M)$
		\STATE $E =  \{C^*_1, C^*_2\}$
		\ENDPROCEDURE\\
	\end{algorithmic}
\end{footnotesize}
\end{algorithm}

\subsubsection{Implicit Context}
In this approach, we utilize mean-pooled embedding of the mention encodings taken from the entire article as an input. Firstly, the entire text is used as an input to obtain the tokenwise encodings from the model.

\begin{align}
	f(text, \theta) = \{E_{T_1}, E_{T_2}, ..., E_{T_n}\}
\end{align}
Here, $E_T$ is encoding of token $T$ and $n$ are the number of tokens in the input text.

Given a span $s$, consisting of $l$ tokens and tokens in the span $\{T_k,....,T_{k+l}\}$, we take the corresponding encodings from the model outputs $\{E_{T_k}, ..., E_{T_{k+l}}\}$. We perform a mean pooling on these encodings to obtain the updated query representation $Q =\frac{1}{l} \sum_{k}^{k+l} \{E_{T_k}, ..., E_{T_{k+l}}\}$. The prototype space consists of the sentence encodings of the canonical names of all the entities in UMLS.

The R@1 candidate generation performance drops drastically in this setup where a drop of more than $30\%$ is observed.  Overall, we observe that these implicit contextual queries are not helpful in improvement of retrieval performance.
\subsubsection{Evaluation}
In this section, we perform the quantitative and qualitative analysis of our context based approaches on the Medmentions st21pv version.  The qualitative examples shown below highlight the predictions provided by the proposed context based approaches. As discussed in the qualitative analysis of region B (see section~\ref{qualitative}), the surface forms have missing context resulting in  an inaccurate prediction. 

It can be observed in table~\ref{table:contextqual} that the mention \textit{mice} is correctly predicted as the entity \textit{\uline{Laboratory mice}} using the MiniEL$_0^{AC}$ and MiniEL$_{NC}$ reranking approaches. We also highlight the effect semantic type reranking approach though the example mentions \textit{kindlin-2} and \textit{kindlin-3} where the prediction semantic type changed from 'Gene' to the correct type 'Protein'. Here, the MiniEL$_0^{NC}$, MiniEL$_0^{AC}$ and MiniEL$_0^{IC}$ methods correspond to the results obtained using the Neighboring Context,  Attention-based Context and Implicit Context approaches, respectively, utilizing MiniEL$_0$ as the base model.

 \begin{table}
	\resizebox{7cm}{!}{\begin{tabular}{p{10cm}}%
			\toprule
	\textbf{\texttt{MENTION}}: \textit{"....iron accumulation in the substantia nigra (SN) of \uline{mice}...."}\\
	\textbf{\texttt{MENTION$^{AC}$}}:  \textit{"...iron accumulation in the substantia nigra (SN) of \uline{mice}\textcolor{brown}{: experiment}....}"\\
	\textbf{\texttt{PREDICTION$^{AC}$}}: \textcolor{blue}{Laboratory mice} (C0025929))\\
	\textbf{\texttt{PREDICTION$^{IC}$}}:  \textcolor{blue}{House mice} (C0025914)\\
	\textbf{\texttt{PREDICTION$^{NC}$}}:  \textcolor{blue}{Laboratory mice} (C0025929)\\
	\midrule
	 \textbf{\texttt{MENTION}}: \textit{ \uline{Kindlin-1} is expressed primarily in epithelial cells,  \uline{kindlin-2} is widely distributed and is particularly abundant in adherent cells, and  \uline{kindlin-3} is expressed primarily in hematopoietic cells.}\\
	\textbf{\texttt{MENTION$^{AC}$}}: \textit{ \uline{Kindlin-1: kind,primarily} is expressed primarily in epithelial cells,  \uline{kindlin-2: distributed,Kind} is widely distributed and is particularly abundant in adherent cells, and  \uline{kindlin-3: expressed,Kind} is expressed primarily in hematopoietic cells.}\\
	\textbf{\texttt{PREDICTION$^{AC, IC, NC}$}}: \textcolor{blue}{FERMT1 gene} (C1423809), \textcolor{blue}{FERMT2 gene} (C1423716), \textcolor{blue}{FERMT3 protein, human} (C1311640)\\
	\textbf{\texttt{PREDICTION$^{AC}$ + TYPE}}: \textcolor{blue}{Fermitin Family Homolog 2, human} (C3889282), \textcolor{blue}{Fermitin Family Homolog 2, human} (C3889282), \textcolor{blue}{FERMT3 protein, human} (C1311640)\\
	\bottomrule
\end{tabular}%
}
\caption{This table shows the qualitative analysis of the MiniEL$_0^{AC}$, MiniEL$_0^{IC}$ and MiniEL$_0^{NC}$ approaches on examples from Medmentions.}
\label{table:contextqual}
\end{table}

It can be observed that the AC approach provides meaningful outputs as it includes the necessary context in the surface form. Similar outputs are provided by the NC approach. However, the neighbouring words may not necessarily contain the context and this can be seen in the following qualitative example listed in the table~\ref{table:ac_nc}.
 \begin{table}[t!]
	\resizebox{7cm}{!}{\begin{tabular}{p{10cm}}%
			\toprule
	 \textbf{\texttt{MENTION}}:"...inhibitor of T cell \uline{function}....hypoxic conditions influence human T cell \uline{functions} and found that..."\\
	\textbf{\texttt{MENTION$^{AC}$}}:"...inhibitor of T cell \uline{function: cell}....hypoxic conditions influence human T cell \uline{functions: cell} and found that..."\\
	\textbf{\texttt{GOLD}}: \textcolor{blue}{Cell physiology} (C0007613), \textcolor{blue}{Cell physiology} (C0007613)\\
	\textbf{\texttt{PREDICTION$^{AC}$}}: \textcolor{blue}{Cell physiology} (C0007613),  \textcolor{blue}{Cell physiology} (C0007613)\\
	\textbf{\texttt{PREDICTION$^{NC}$}}: \textcolor{blue}{Cell physiology} (C0007613),	 \textcolor{blue}{T cell differentiation} (C1155013)\\
	\bottomrule
\end{tabular}%
}
\caption{This table shows the qualitative analysis of the MiniEL$_0^{AC}$ and MiniEL$_0^{NC}$ approaches on examples from Medmentions.}
\label{table:ac_nc}
\end{table}

\begin{table}[htpb!]
	\centering
	\renewcommand{\arraystretch}{1.3}
	\resizebox{.25\textwidth}{!}{% <------ Don't forget this %
	\begin{tabular}{lrr}
		\toprule 
		Approach & R@1 & R@5 \\
		\midrule
		miniEL$_0$ & 0.553&0.756\\
		miniEL$_0^{NC}$ & 0.219 & 0.405  \\
		miniEL$_0^{AC}$ & 0.384 & 0.642  \\
		miniEL$_0^{IC}$ & 0.161 & 0.359\\
		\bottomrule
	\end{tabular}%
}
	\caption{The table presents the candidate generation performance of the listed context based approaches. The performance is computed the st21pv version of Medmentions.}
	\label{tab:contextualresults}
\end{table}

\begin{figure}[htpb!]
	\centering
	\resizebox{8cm}{!}{\includegraphics[width=\textwidth, height=0.8\textwidth, keepaspectratio]{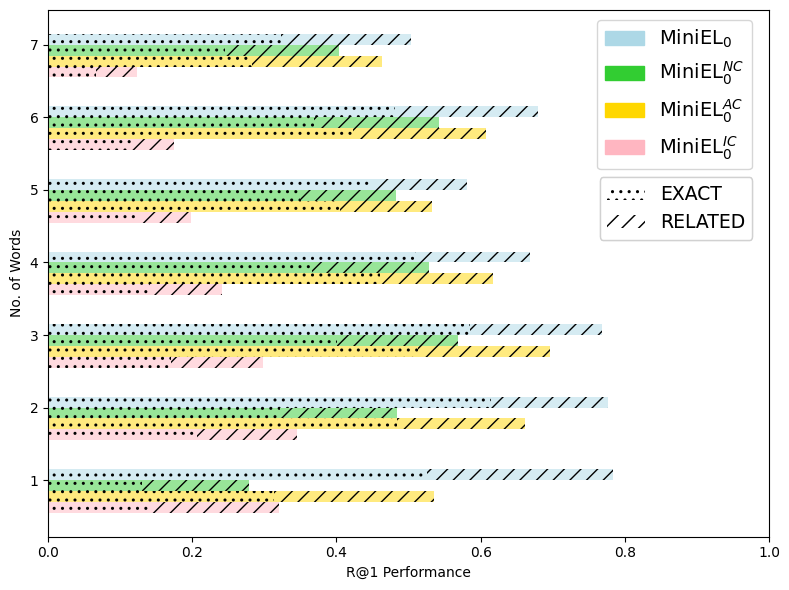}}
	\caption{This figure presents the word-count level retrieval performance, measured in terms of exact and related matches, comparing the performance of the MiniEL$_0$ approach in comparison to its performance on applying  the context based methods.}
	\label{fig:contextanalysis}
\end{figure}

We observe that the attention span based context enrichment approach is sensitive to the context addition as it induces bias the surface form and the resulting candidates may be more similar to the bias term as compared to the base form. Therefore, to understand the impact of bias on the surface form, we observe the retrieval performance based on the number of words in the mention. The figure~\ref{fig:contextanalysis} shows that the performance of MiniEL$_0^{AC}$ is better on mentions with higher length as compared to the 
mentions with lower lengths. A similar trend is observed for the MiniEL$_0^{NC}$ approach. This trend is not seen for the MiniEL$_0^{IC}$ approach where the performance drops with the increase in number of words in the mentions. However, the attention span based approach has better performance as compared to the neighboring context approach. For each specific mention word count, we select mentions with at least about a 100 examples for this analysis.

To summarize, the quantitative and qualitative context enrichment analysis shows that the MiniEL$_0^{AC}$ approach outperforms the other approaches and is effective in context addition. However, the sensitivity in the encodings results in large deviations in the candidate generation (see table~\ref{tab:contextualresults}). Therefore, the robustness of this contextual approach needs to be improved.

\end{document}